\documentclass[runningheads]{llncs}

\usepackage{eccv}

\usepackage{eccvabbrv}

\usepackage{graphicx}
\usepackage{booktabs}

\usepackage[accsupp]{axessibility}  % Improves PDF readability for those with disabilities.
\usepackage{adjustbox}

\usepackage[pagebackref,breaklinks,colorlinks,citecolor=eccvblue]{hyperref}

\usepackage{orcidlink}

\makeatother
\usepackage{xcolor}
\definecolor{lightpurple}{rgb}{0.7, 0.3, 0.8}

\usepackage{lipsum}

\newcommand\blfootnote[1]{%
  \begingroup
  \renewcommand\thefootnote{}\footnote{#1}%
  \addtocounter{footnote}{-1}%
  \endgroup
}
\begin{document}

% ---------------------------------------------------------------
% TODO REVIEW: Replace with your title
\title{STAG4D: Spatial-Temporal Anchored Generative 4D Gaussians} 

% TODO REVIEW: If the paper title is too long for the running head, you can set
% an abbreviated paper title here. If not, comment out.
\titlerunning{STAG4D}

% TODO FINAL: Replace with your author list. 
% Include the authors' OCRID for the camera-ready version, if at all possible.

\author{Yifei Zeng\inst{1}* \and
Yanqin Jiang\inst{2}* \and
Siyu Zhu\inst{3} \and
Yuanxun Lu\inst{1} \and
Youtian Lin\inst{1} \and
Hao Zhu\inst{1} \and
Weiming Hu\inst{2} \and
Xun Cao\inst{1} \and
Yao Yao\inst{1} 
}

%\cortext[equal]{Equal Contribution.}
%\cortext[equalcontrib]{Equal Contribution.}
%
%% TODO FINAL: Replace with an abbreviated list of authors.
\authorrunning{Yifei Zeng et al.}
%% First names are abbreviated in the running head.
%% If there are more than two authors, 'et al.' is used.
%
%% TODO FINAL: Replace with your institution list.
\institute{ Nanjing University \and
 Institution of Automation, Chinese Academy of Science \and
 Fudan University}
%\email{lncs@springer.com}\\
%\url{http://www.springer.com/gp/computer-science/lncs} \and
%ABC Institute, Rupert-Karls-University Heidelberg, Heidelberg, Germany\\
%\email{\{abc,lncs\}@uni-heidelberg.de}}

\maketitle

\begin{abstract}
Recent progress in pre-trained diffusion models and 3D generation have spurred interest in 4D content creation. However, achieving high-fidelity 4D generation with spatial-temporal consistency remains a challenge. 
In this work, we propose STAG4D, a novel framework that combines pre-trained diffusion models with dynamic 3D Gaussian splatting for high-fidelity 4D generation. 
Drawing inspiration from 3D generation techniques, we utilize a multi-view diffusion model to initialize multi-view images anchoring on the input video frames, where the video can be either real-world captured or generated by a video diffusion model. 
To ensure the temporal consistency of the multi-view sequence initialization, we introduce a simple yet effective fusion strategy to leverage the first frame as a temporal anchor in the self-attention computation. 
With the almost consistent multi-view sequences 
%as spatial and temporal anchors
, we then apply the score distillation sampling to optimize the 4D Gaussian point cloud.
The 4D Gaussian spatting is specially crafted for the generation task, where an adaptive densification strategy is proposed to mitigate the unstable Gaussian gradient for robust optimization. 
Notably, the proposed pipeline does not require any pre-training or fine-tuning of diffusion networks, offering a more accessible and practical solution for the 4D generation task.
Extensive experiments demonstrate that our method outperforms prior 4D generation works in rendering quality, spatial-temporal consistency, and generation robustness, setting a new state-of-the-art for 4D generation from diverse inputs, including text, image, and video. Project Page: \url{https://nju-3dv.github.io/projects/STAG4D}
\blfootnote{*Equal Contribution}

% 
% Drawing from 3D generation techniques, we utilize a multi-view diffusion model to initialize multi-view images based on a single image input. Spatial consistency is achieved through cross-view attention, which is anchored to a reference frame.

% For temporal consistency, our approach takes a coherent video as input and introduces a simple yet effective temporal fusion strategy to leverage the first frame as a temporal anchor for cross-frame attention computation, which guides the coherent multi-view generation of subsequent frames. With the coherent multi-view images at all frames, we then apply the multi-view score distillation sampling loss for optimization of the 4D Gaussian points.


\end{abstract}

\section{Introduction}
\label{sec:intro}

Recent advancements in large-scale pre-trained diffusion models have shown remarkable progress in producing high-quality and diverse visual content, including images, videos, and 3D assets~\cite{poole2022dreamfusion, rombach2022high, ho2022imagen, wang2023modelscope, 2023i2vgenxl, liu2023zero, shi2023MVDream,lu2023direct2}. 
The progress naturally extends to the realm of dynamic 3D generation~\cite{singer2023text4d, jiang2023consistent4d, zheng2023unified, bahmani20234d, ling2023align, yin20234dgen, zhao2023animate124, wang2023animatabledreamer}. This endeavor has gained prominence in computer vision and generative AI research, as high-quality 4D content generation is key to a broad range of applications such as autonomous driving simulation, game and film industries, digital Avatar creation, and spatial video production.
However, previous 4D generation methods face challenges including blurry rendering, spatial-temporal inconsistency, and slow generation speed. Generating high-quality 4D content efficiently and practically remains a significant challenge.

\begin{figure*}[t]
  \centering
  \includegraphics[width=1.0\linewidth]{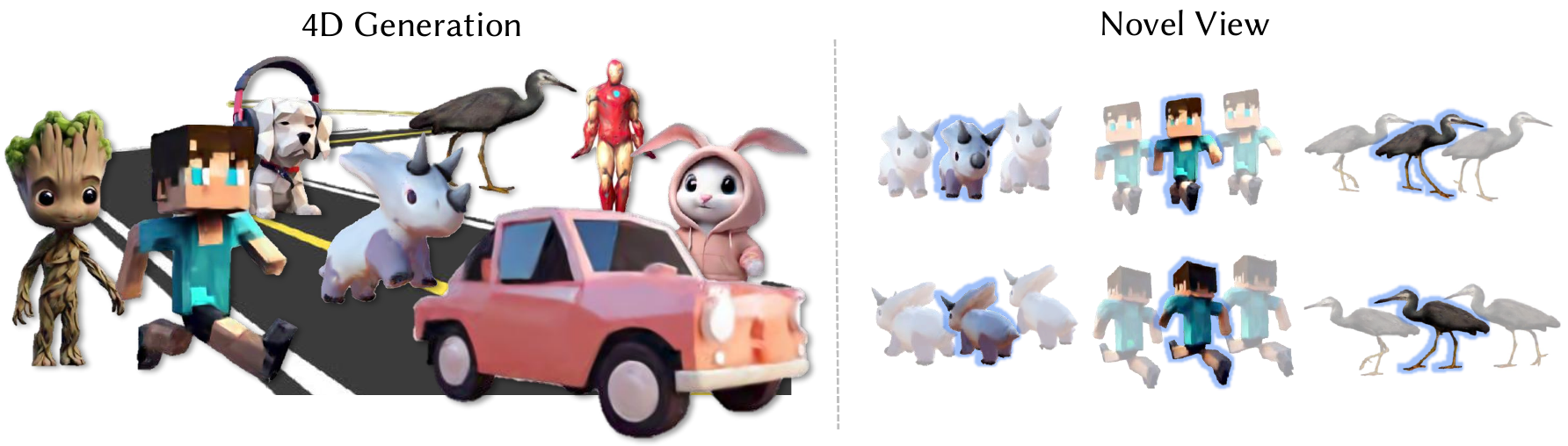}
  \vspace{-0.3in}
  \caption{Visualization of the generated 4D asserts. Our approach can generate diverse 4D content from various inputs, including text, image, and video.}
  \vspace{-0.1in}
\label{fig:teaser}
\end{figure*}

Given the promising strides in 3D generation from pre-trained diffusion models, there has been a concerted focus on generalized dynamic 3D generation from text or uncalibrated monocular video~\cite{singer2023text4d, ling2023align, bahmani20234dfy, zheng2023unified, jiang2023consistent4d, wang2023animatabledreamer}. 
MAV3D~\cite{singer2023text4d} is the pioneering work of text-to-4D generation which creates 4D scenes from textual descriptions through Score Distillation Sampling (SDS) from a video diffusion model. 
% Subsequent text-to-4d works~\cite{zheng2023unified, bahmani20234dfy, ling2023align, zhao2023animate124} followed its paradigm.
Alternatively, Consistent4D~\cite{jiang2023consistent4d} takes a video as input, and focuses on the task of video-to-4D generation. The pipeline is more flexible in that it can take advantage of high-quality input videos, dividing the difficult problem of realistic 4D content creation into a practical pipeline of text-to-video and video-to-4D generation.
However, current video-to-4D generation methods still struggle to generate high-fidelity 4D asserts with spatial and temporal rendering consistency. Meanwhile, the commonly used Neural Radiance Field(NeRF) representation would suffer from an over-saturated appearance and a long optimization time.

In this paper, we address two important factors in 4D content creation: a suitable 4D representation and spatial-temporal consistency imposed during pseudo-projection generation (i.e., multi-view video generation). 
Our approach employs a 4D Gaussian splatting approach specially tailored for the generation task. 
In the generation process, we first generate the multi-view images at all timestamps of the 4D scene.
To enforce the 4D consistency of the multi-view video initialization, we design a direct fusion approach for attention computation in both spatial and temporal domains. This approach implicitly addresses texture degradation and geometric misalignment, thereby avoiding the need for adding an explicit multi-view or temporal consistency loss during the Score Distillation Sampling(SDS)~\cite{poole2022dreamfusion} optimization.
Then, the reference photometric loss and the SDS loss conditioned on the initial multi-view video are applied for 4D Gaussian point cloud optimization. 
To tackle the instability in optimizing the 4D Gaussian points, we introduce an adaptive densification strategy that is informed by the Gaussian gradient distribution. This approach yields high-quality and robust 4D scene generation from a monocular video input.

We have conducted comprehensive experiments to demonstrate the effectiveness of each component of the proposed method. Our approach achieves a 2x faster generation speed compared with the previous video-to-4D approaches (e.g.,~\cite{jiang2023consistent4d}) and a significantly better generation quality than prior state-of-the-art methods. It is also noteworthy that the 4D content generated by our method can be rendered in real-time, opening up a wide range of possibilities for practical applications. In summary, our primary contributions are as follows:

\begin{itemize}

\item We introduce a holistic 4D generation pipeline that streamlines the generation process into sequential stages: video generation, multi-view video initialization, and 4D optimization using multi-view video conditioned SDS.

\item We harness the dynamic 3D Gaussian representation for 4D generation, complemented by an adaptive densification strategy. This combination enables highly precise and efficient 4D generation from monocular video inputs.

\item We present a novel training-free attention fusion module to effectively integrate temporal anchor frames into the multi-view diffusion process, significantly enhancing the 4D consistency of the generated multi-view videos.

\item Our method outperforms previous approaches in optimization efficiency, rendering quality, and 4D consistency, establishing a new benchmark for 4D generation across various input types, such as text, images, and videos.
    
\end{itemize}

% contributions
% We have performed comprehensive experiments on both public and private data. To summarize, our main contributions are:   \begin{itemize} 
%     \item We propose a novel framework for 4D dynamic object generation by adapting 4D gaussian to generative settings, which significantly reduces the training time compared to previous work. 
%     \item We design TimeMix, a training-free temporal attention module, and obtain a temporally consistent supervision signal from image diffusion model with minimal negative impact on image quality and 3D consistency. % With this simple operation, we achieve remarkable results. 
%     \item We propose an adaptive densification strategy for 4D gaussian based on object dynamics, which enhances the generation quality and robustness of dynamic regions considerably. % compared with the conventional densification strategy. 
%     \item We demonstrate the effectiveness of our method with extensive results on public and private data. Our method can generate a high-quality 4D dynamic object from a 2-second video within 30 minutes on a single 3090 GPU. 
% \end{itemize}

\section{Related work}
\subsection{3D Generation}

% Recent advancements in the generative modeling of static 3D content have garnered considerable attention. 
% Existing 3D generation algorithms can be categorized into two primary groups: optimization-based and learning-based methods.
% In the context of Neural Radiance Fields (NeRF), optimization-based methods refine an explicit 3D representation through score distillation sampling derived from an image diffusion model. 
% These methods typically require extensive computational time, often exceeding tens of minutes, to generate a single 3D object of high quality.
% On the other hand, learning-based methods are capable of generating 3D objects in a matter of minutes or even seconds during inference. 
% Despite their efficiency, these methods necessitate pre-training on extensive datasets of 3D objects and are subject to biases inherent in the training data.

Dreamfusion~\cite{poole2022dreamfusion} first proposes the SDS loss~\cite{poole2022dreamfusion,wang2023score} for NeRF optimization, which stands for the most prevalent technique for nowadays text-to-3D approaches. To mitigate the multi-view Janus problem, Zero123~\cite{liu2023zero} and SycnDreammer~\cite{liu2023syncdreamer} fine-tune 2D diffusion models to grant the image generator with the ability of viewpoint control, while later works~\cite{shi2023MVDream, shi2023zero123++, long2023wonder3d, lu2023direct25} explicitly generate fixed multi-view images in one diffusion pass. Building upon the success of multi-view diffusion methods, we integrate a multi-view generation module into the 4D generation task, with meticulous handling of 4D spatial-temporal consistency for enhanced 4D generation quality.

Except for equipping diffusion models with multi-view or depth awareness, another direction of progress focuses on 3D representations. Following the idea of differentiable optimization, Magic3D~\cite{lin2023magic3d} applies a two-stage generation pipeline and changes the 3D representation to instant-NGP~\cite{mueller2022instant} and DMTet~\cite{shen2021dmtet}, achieving faster runtime and better generation quality. Direct2.5~\cite{lu2023direct25} applies an explicit mesh representation and uses differentiable rasterization for fast mesh optimization. Recently, with the newly developed 3DGS, DreamGaussian~\cite{tang2023dreamgaussian} demonstrates the ability to generate 3D objects within several minutes by substituting NeRF with the 3D Gaussian representation, showing the potential of using an explicit point-based representation in 3D generation tasks. In this work, we introduce a novel 4D Gaussian representation and a specially tailored optimization scheme for the 4D generation task.

% These enhancements are specifically designed to reduce the necessity for extensive fine-tuning when utilizing pre-trained image diffusion models, such as Stable Diffusion.
% We extend this approach by incorporating a multi-view consistent diffusion process proposed in Zero123++ into the temporal domain, aiming to achieve a more coherent and temporally consistent generation of dynamic scenes.

% general
% Recently, exciting progress has been made in 3D generation area.
% category by generation process
% Existing 3D generation algorithms can be categoried to two groups: optimization-based and learning-based.
% optimization-based
% Optimization-based methods optimize a 3D representation, e.g., NeRF, via SDS from an image diffusion model.
% They usually take more than 30 minutes to generate a 3D object whilst enjoy high generated quality.
% On exception is DreamGaussian, which only takes two minutes to generate a 3D object by replacing NeRF with 3D Gaussian Splattingm a much more efficient 3D representation.
% learning-based
% On the other hand, learning-based methods usually generate 3D objects in minutes or even seconds when inference. 
% But they require pre-training on large-scale 3D object datasets, and inevitably suffer from bias in training data.
% relate to our work
% The above 3D generation works inspire 4D generation works, and our method for efficiency training is also inspired by one of 3D generation method, DreamGaussian.

\subsection{4D Reconstruction}

Dynamic 3D reconstruction is yet another heated research topic in the field of computer vision and graphics. 
By extending the static NeRF~\cite{mildenhall2020nerf} framework to dynamic scenes, dynamic NeRFs~\cite{nsff,xian2020space,gao2021dynamic,du2021neural} have demonstrated remarkable progress on dynamic 3D reconstruction. However, limited by the implicit neural representation and the complex nature of dynamic 3D information, dynamic NeRFs still suffer from slow optimization speed and low reconstruction quality.
With the development of 3DGS~\cite{kerbl20233d}, researchers have quickly extended the 3D Gaussian to dynamic scene representation, which achieves faster training and rendering speeds compared to D-NeRFs based on implicit neural representations. Dynamic 3D Gaussian~\cite{luiten2023dynamic} applies the per-frame 3DGS optimization for 4D scene reconstruction. Some studies~\cite{wu20234d,yang2023deformable3dgs} attempt to amalgamate the explicit point-based 3DGS with an implicit neural field for dynamic information modeling, while Gaussian-Flow~\cite{lin2023gaussian} introduces an explicit per-point motion model to represent a 4D scene without using implicit neural networks. It is noteworthy that it is non-trivial to extend the Gaussian reconstruction to 4D reconstruction as each reconstruction pipeline requires heuristic tuning of hyper-parameters at each stage. In this paper, we target combining the 4DGS for the challenging task of 4D generation.   

% general
% 4D reconstruction, aka dynamic scenes reconstruction, is a long-standing problem.
% category by representation
% Various 4D representations are utilized for reconstruction, including point cloud, mesh, dynamic NeRF and dynamic Gaussian.
% dynamic NeRF
% Dynamic NeRF, an extension of NeRF to dynamic scenes, has been well studied in the last few years and achieve remarkable progress in both rendering quality and training speed.
% However, it still has to do trade-offs between quality and speed.
% 4D gaussian
% In contrast, recently, researchers extent 3D Gaussian Splatting to dynamic scenes and achieve competitive quality when compared with dynamic NeRFs, whilst enjoying faster training and rendering speed.
% TODO: check it 
% For example, dy3D, the first dynamic Gaussian, proposes to first reconstruct a static scene and then learn the per-frame Gaussian position, roation, traslation and scale in future frames.
% Following works including xx, and xx, both exploit a deformation field to better predict the offset and appearance change of the gaussian points in the whole sequence.
% realte to our work
% Given the superior performance of xx in reconstruction task, we decide to introduce it to generation task in this work.

\subsection{4D Generation}

Recent developments in 3D content generation and video diffusion technologies have triggered researchers' interest in exploring 4D content generation from various input conditionings. MAV3D~\cite{singer2023text4d} presents the early attempt at text-to-4D generation. By utilizing score distillation sampling derived from video diffusion models, MAV3D optimizes a dynamic NeRF based on textual prompts. In a parallel vein, Consistent4D~\cite{jiang2023consistent4d} has introduced the task of video-to-4D. This method capitalizes on the pre-trained knowledge from image diffusion models to optimize dynamic NeRFs via SDS optimization, showcasing the potential of high-quality 4D content generation from pre-trained 2D diffusion models. Concurrently with our work, 4DGen~\cite{yin20234dgen} has introduced a framework for generating dynamic 3D models, which employs spatial-temporal pseudo labels on keyframes within a multi-view diffusion model. However, the overall quality of its generation can be further improved, as its rendering fidelity is limited by the implicit neural representation and inadequate spatial-temporal information exchange exists in the diffusion generation process.
In contrast, our method applies an efficient 4D Gaussian representation with spatial-temporal attention from reference- and the first-frame anchors, achieving substantially better generation results than previous works.
% \jiang{TODO: add “all” text-to-4d generation works, please refer to my statements in intro. }
% general
% TODO: check it 
% Inspired by advances in 3D generation, some pioneering works make attempts in 4D generation~\cite{singer2023text, jiang2023consistent4d, shao2023control4d, liu2023dynvideoe}.
% MAV3D~\cite{singer2023text} is the first text-to-4d method. It adopts SDS from video diffusion model to optimize a dynamic NeRF given a text prompt as the input, sharing similar pipeline as DreamFusion~\cite{poole2022dreamfusion}.
% video-to-4d
% Recently, Consistent4D~\cite{jiang2023consistent4d} proposes a new task of generating 360-degree 4D dynamic object from uncalibrated monocular video, short as video-to-4D. A 3D-aware image diffusion model~\cite{liu2023zero} is adopted to provide main supervision signals for the optimization of dynamic NeRF via SDS. The authors claim spatial and temporal consistency as the main challenge of this task and they propose an interpolation-driven consistency loss to alleviate the inconsistency.
% relate to ours
% However, Consistent4D suffer from long training time, and thus in this work, we speed it up without sacrificing generation quality by refining crucial designs in their framework.

% \input{sec/preliminary}
\section{Method}
\label{sec: method}

\begin{figure*}[t]
  \centering
  \includegraphics[width=1.0\linewidth]{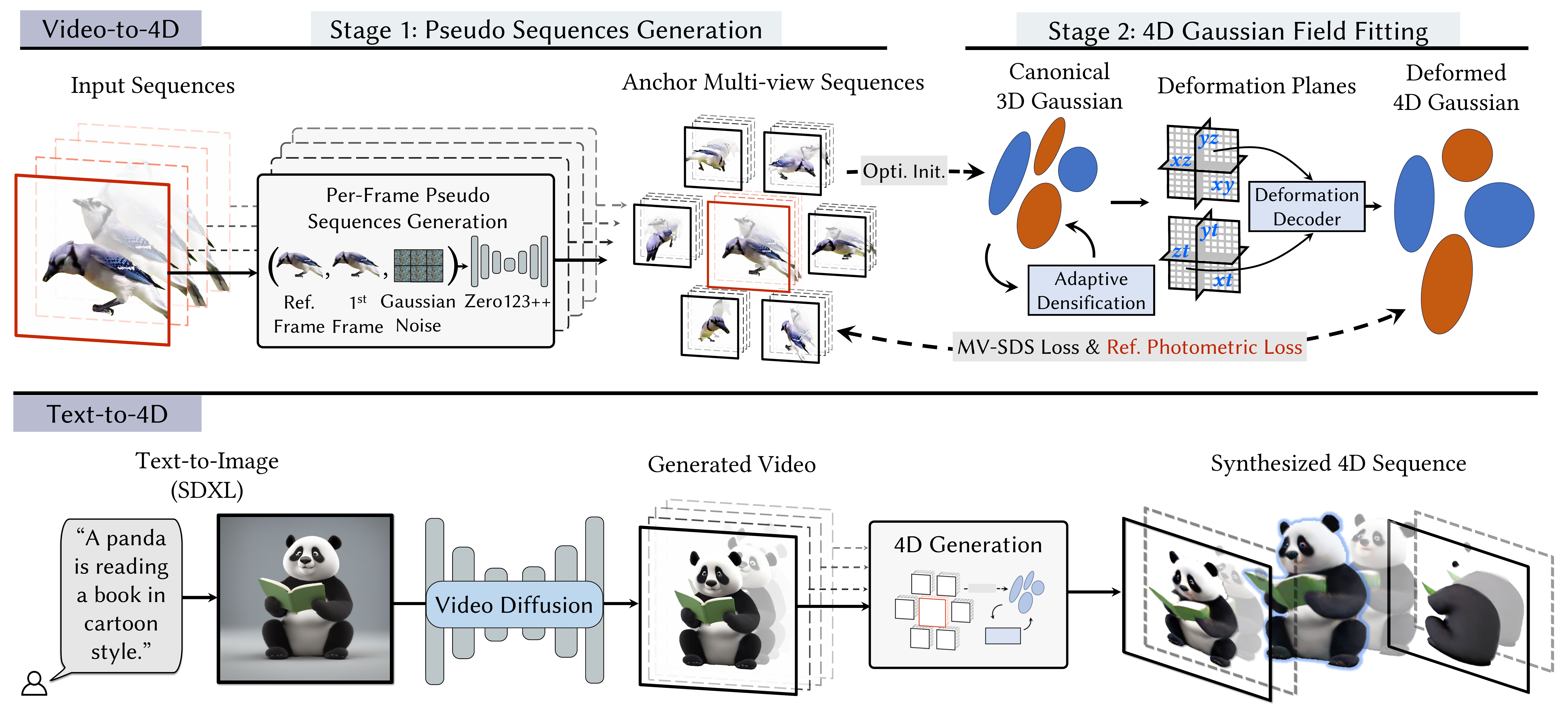}
  \vspace{-0.2in}
  \caption{\textbf{Overall pipeline.} Given a video input, we apply a multi-view diffusion model to produce coherent multi-view sequences, which serve as spatial and temporal anchors. Next, we train a deformable 3D Gaussian using multi-view SDS loss and reference loss. For a text-to-4D generation, our pipeline can be naturally extended to accept the text input by integrating with an off-the-shelf text-to-video module. }
  \vspace{-0.1in}
\label{fig:pipeline}
\end{figure*}
The framework of the proposed approach, as depicted in Figure~\ref{fig:pipeline}, outlines the generation and optimization process of a dynamic scene from a given video input. Also, our approach can be easily extended to a text-to-4D generation pipeline by utilizing an off-the-shelf video generation module.
Section~\ref{subsec:4drepresentation} presents the 4D representation used in the proposed method, along with an adaptive densification strategy that adjusts the densification threshold based on the relative motion gradient of the scene points.
Subsequently, Section~\ref{subsec: spatial_temporal_consistency} provides a brief overview of the multi-view diffusion model used in the pipeline and introduces a novel spatial-temporal attention fusion mechanism designed to produce the almost consistent multi-view image sequences.
Finally, as detailed in Section~\ref{subsec:training_objectives}, the generated multi-view sequences are leveraged with spatial and temporal anchor frames for consistent 4D generation with the SDS optimization.

\subsection{4D Representation}
\label{subsec:4drepresentation}
\subsubsection{4D Gaussian Splatting}
The concept of 3D Gaussian Splatting was initially introduced in the work of~\cite{kerbl20233d}, where an explicit point-based representation is used to model the 3D scene. Specifically, the 3D Gaussian point cloud, denoted as $\mathcal{S}$ for a static scene, is characterized by the tuple:
\begin{equation}
\mathcal{S} = [\mathcal{X}, s, r, \sigma, \zeta],
\end{equation}
where $\mathcal{X} = (x, y, z)$ denotes the positions of the 3D Gaussian points, and $s$, $r$, $\sigma$ and $\zeta$ represent the scale, rotation, opacity, and spherical harmonics (SH) coefficients of the radiance, respectively. In 3D reconstruction, the 3D Gaussian point cloud will be optimized through a point-based differentiable volume rendering.

This representation can be naturally extended to 4D scenes by presenting the continuous scene dynamics as a 3D motion field. Inspired by~\cite{wu20234d}, we present a 4D scene as a 3D Gaussian point cloud with a hex-plane-based deformation field, denoted as $\mathcal{F}(\mathcal{S}, t)$. The 3D Gaussian point cloud at time $t$ can be expressed as:
\begin{equation} 
\mathcal{F}(\mathcal{S}, t) = [\mathcal{X}_t, s_t, r_t, \sigma, \zeta], 
\end{equation}
where $\mathcal{X}_t = (x_t, y_t, z_t)$, $s_t$ and $r_t$ represent the updated information of Gaussian position, scale, and rotation at time $t$.

% Dynamic gaussian splatting present high quality novel view synthesis and fast training and rendering speed.
%
% Inspired by recent work~\cite{wu20234d}, we use 4D Gaussian to represent our dynamic scene. 4D Gaussian first learns a 3D Gaussian $\mathcal{S} = [\mathcal{X}, s, r, \sigma, \zeta]$ for static scene, where $\mathcal{X}$ represents the position of 3D Gaussian points, $s$ and $r$ denote the scale and rotation of the Gaussian points, and $\sigma$ and $\zeta$ are opacity and SH coefficients respectively. Then they train a deformation field $\mathcal{F}$, represented by hex-planes, to predict the 3D Gaussian information $\mathcal{S}'$ at time $t$. Specifically, we have:
% \begin{equation}
%    \mathcal{S'} = [\mathcal{X}', s', r', \sigma, \zeta] = \mathcal{F}(\mathcal{S},t),
% \end{equation}
% where $\mathcal{X}'$, $s'$ and $r'$ represents the updated information of gaussian position, scale and rotation at time $t$.
% densify
% Note that 4D Gaussian adopts the densification strategy in 3D gaussian, i.e., a fixed threshold of view-space position gradient. 
% While the fixed threshold may be suitable under reconstruction settings, our experiment shows it performs bad in generative settings, so we opt for an adaptive threshold, as discussed in Sec.~\ref{sec: method-adaptive-densify}.

\subsubsection{Adaptive Densification}
The standard 3D Gaussian Splatting technique typically employs a point cloud densification control strategy to dynamically adjust the number of Gaussians and their density within a unit volume. 
This adaptive approach allows for the transition from an initial sparse Gaussian set to a denser configuration to better represent the 3D scene. 
The 4D Gaussian Splatting method introduced in the work~\cite{wu20234d} utilizes a densification strategy similar to that presented in vanilla 3DGS~\cite{kerbl20233d}, which involves the use of a fixed densification threshold based on the view-space position gradient.
However, while the fixed gradient threshold strategy demonstrates efficacy in the context of reconstruction, particularly when multiple views afford robust and redundant coverage of the target scene, it performs sub-optimally in the generative setting. 
This limitation arises from the constraints imposed by the input single image or monocular videos, leading to significant uncertainty in the spatial and scale dimensions for each training object. This will lead to different optimal thresholds for different cases, as illustrated in the ablation study.
% \yifei{maybe need to modify from here}

% To address this issue, we propose an adaptive threshold approach, wherein only candidates with relatively large gradients are selected for densification. 
% Furthermore, we maintain a constant proportion of points to be densified during each operation, resulting in a smoother increase in point numbers and facilitating the convergence of the deform training process.
% Experimental results demonstrate that the proposed adaptive densification strategy significantly enhances the generation quality and robustness of dynamic scenes.
% \yifei{writing the new draft for adaptive densification here.}  

To address the issue,  we propose an adaptive threshold approach, wherein only candidates with relatively large gradients are selected for densification. Our approach is based on the statistical analysis of the accumulated gradient of each point, which follows a log distribution of similar shapes throughout the training process.
We apply an adaptive threshold that filters out Gaussians with a relatively small gradient and only densifies those with a large gradient. The adaptive threshold is set to select a fixed percentage (top $\lambda\%$) of points with the highest gradient in each densification operation.
This ensures that the threshold adapts to the distribution of the gradient and maintains a stable relative position. Experimental results demonstrate that the proposed simple strategy can significantly enhance the quality and robustness of the 4D generation.

\subsection{Temporal and Multi-view Consistent Diffusion}
\label{subsec: spatial_temporal_consistency}
\subsubsection{Multi-view Consistent Diffusion}
% general intro
Our pipeline adopts Score Distillation Sampling(SDS)~\cite{poole2022dreamfusion} from image diffusion models for the optimization of dynamic Gaussian. Specifically. Image-to-image diffusion model Zero123~\cite{liu2023zero} is utilized and the SDS loss gradient could be formulated as follows: 
% \begin{equation}\label{eq:sds}
% % \begin{aligned}
%     \nabla_{\theta} \mathcal{L}_{SDS}(\phi, \mathbf{x}) = \mathbb{E}_{t, \epsilon} \left [ \omega(t)(\hat{\epsilon}_\theta(\mathbf{z}_t;\mathbf{I}_{in}, t) - \epsilon) \frac{\partial \mathbf{x}}{\partial \theta}\right ], \\
%     % \hat{z} &= z_t-\sigma_t\hat{\epsilon}_t(z;\mathbf{I}_{in}).
% % \end{aligned}
% \end{equation}
\begin{equation}
\begin{aligned}
    \nabla_{\theta} \mathcal{L}_{SDS}(\phi, \mathbf{x}) &= \mathbb{E}_{t, \epsilon} \left [ \omega(t)(\hat{\epsilon}_\theta(\mathbf{z}_t;\mathbf{I}_{in}, \mathbf{R}, \mathbf{T}, t) - \epsilon) \frac{\partial \mathbf{x}}{\partial \theta}\right ], \\
    \hat{z} &= z_t-\sigma_t\hat{\epsilon}_t(z;\mathbf{I}_{in}, \mathbf{R}, \mathbf{T}).
\end{aligned}
\end{equation}
where $\theta$ represents the parameters of the 3D representation, $\mathbf{x}$ the rendered image at the current view, $t$ the timestamp in the diffusion process, $\epsilon$ the ground truth noise, $\hat\epsilon$ the predicted noise from the noisy image $\mathbf{z_t}$ conditioned on an initial input $\mathbf{I}_{in}$, and the relative camera pose between input view and target view $(\mathbf{R}, \mathbf{T})$. Zero123 can generate the target view image at any relative camera location but only one target view at a time, thus good at optimizing the 3D object from all views whilst bad at generating spatially consistent images from multiple target views. In contrast, Zero123++~\cite{shi2023zero123++} leverages reference attention~\cite{2023reference} to model the relationship of the images from multiple target views as well as the input view, resulting in multi-view consistency output yet at the cost of fixed target camera locations. 
We combine the advantages of both models by taking the multi-view consistent output of Zero123++ as the input images of Zero123 when calculating SDS loss.

\subsubsection{Temporally Consistent Diffusion}\label{sec: method-timemix}
Following the view generation philosophy of multi-view diffusion based 3D generation, a practical solution to 4D content creation is to generate the multi-view videos of the 4D scene, and then apply the multi-view video conditioned SDS loss to optimize the scene. However, it is rather difficult to generate temporally consistent multi-view videos by the separate per-frame generation. Inspired by zero-shot video generation method~\cite{text2video-zero}, we propose a training-free temporal attention module to enable Zero123++~\cite{shi2023zero123++} with temporal-awareness.

In our design, we obtain the attention information during denoising the multiview images of the first frame. Then we apply the recorded attention to the later denoising process. Specifically, during the denoising process of the latent code, the self-attention layer computes three essential components—queries ($Q$), keys ($K$), and values ($V$), which is formulated as: 
 \begin{equation}
     Self\text{-}Attn(Q,K,V)=Softmax(\frac{QK^{T}}{\sqrt{c}})V.
 \end{equation}
We first compute the key $K_0$ and value $V_0$ for the initial frame, $T_0$.
Subsequently, for each subsequent frame $T_{t}$, where $t\in\{1,...,N\}$, we perform a mixing operation on the key $K_{t}$ and value $V_t$ with the initial key $K_0$ and value $V_0$ derived from the first frame $T_0$. 
 This process can be formally represented as follows:
 \begin{equation}\label{eq:spatial_temporal}
     \begin{cases}
         K_t = \gamma \mathbf{K}_0 + (1-\gamma) \mathbf{K}_t\\
         V_t = \gamma \mathbf{V}_0 + (1-\gamma) \mathbf{V}_t,
     \end{cases}
 \end{equation}
 where the parameter $\gamma$ serves as a weighting factor that governs the influence of the initial frame, as illustrated in the ablation part. Additionally, we employ reference attention following the approach outlined in ~\cite{shi2023zero123++,reference-attention}, to extract local conditioning from the input image. In this reference attention mechanism, both the key and the value are obtained through the concatenation of features from the reference image and the noisy input image. Our design of temporal attention module is simple yet effective with superior temporal consistency, image quality and 3D consistency to vanilla cross-frame attention proposed in~\cite{text2video-zero}, as demonstrated in experiment section.

\begin{figure}[t]
  \centering
  \includegraphics[width=1.0\linewidth]{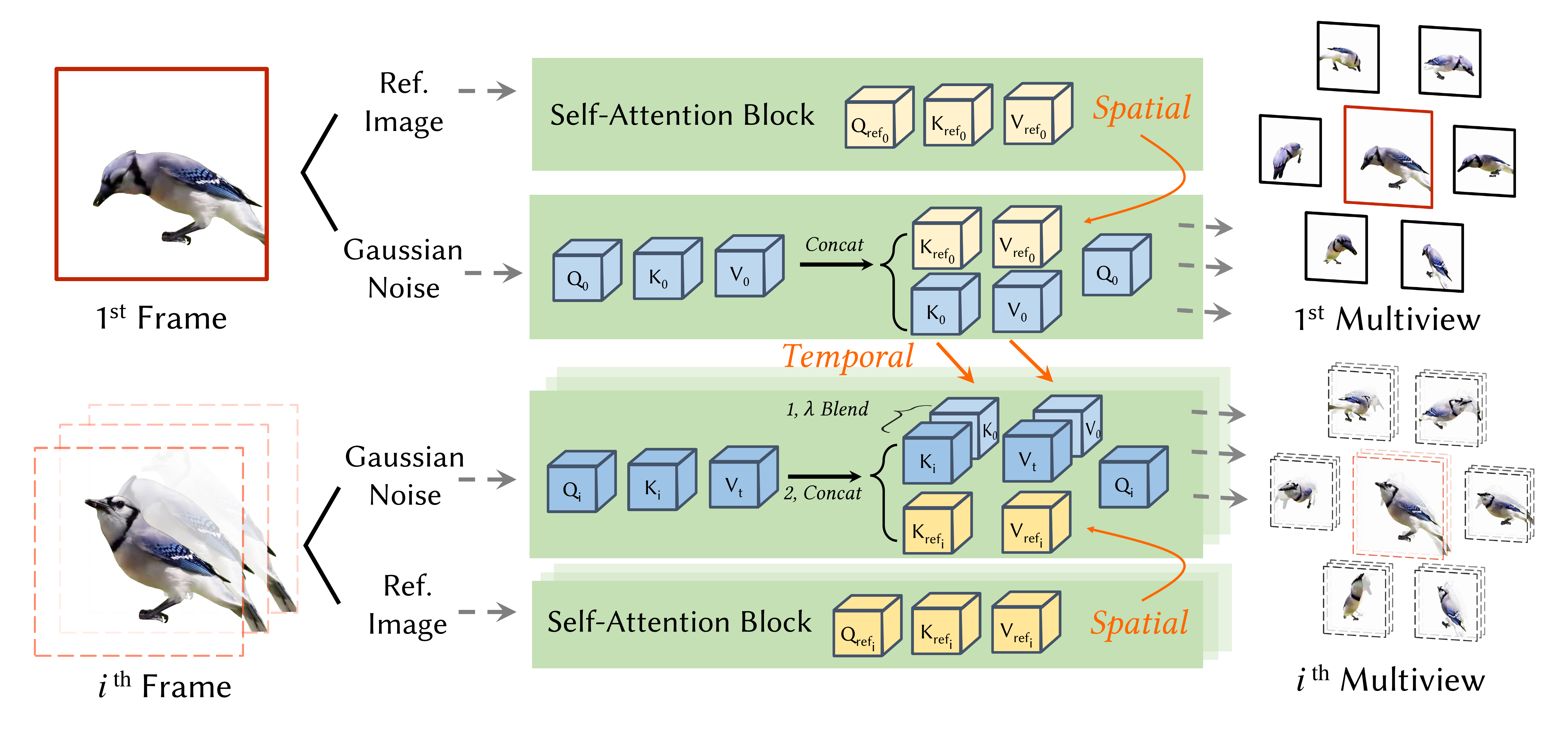}
  %\vspace{-0.3in}
  \caption{Illustration of the proposed spatial and temporal attention fusion during multi-view sequence inference.}  
  \vspace{-0.1in}
\label{fig:time_mix}
\end{figure}

\subsection{Training Objectives}
\label{subsec:training_objectives}
% definition
Following the generation of multi-view sequences from a monocular reference video, we can acquire 6 anchor views $\{I^i_t\}_{i\in\{1...6\}}$ and a reference view $I^{ref}_t$ at each timestep $t$. 
In our optimization process, we employ multi-view score distillation sampling (SDS) utilizing the generated images $\{I^i_t\}_{i=1...6}$ in conjunction with the reference images $I^{ref}_t$.
The multi-view score distillation loss function $\mathcal{L}_{MVSDS}$ can be defined as:
\begin{equation}
    \begin{aligned}
        \mathcal{L}_{MVSDS} & = \lambda_1 \mathcal{L}_{SDS}^{i} + \lambda_2 \mathcal{L}_{SDS}^{ref}\\
         & =\lambda_1 \mathcal{L}_{SDS}(\phi,I^i_t) + \lambda_2 \mathcal{L}_{SDS}(\phi,I^{ref}_t),
    \end{aligned}
\end{equation}
where $\lambda_1$ and $\lambda_2$ are two weighting factors, the index $i$ is determined based on the proximity of the rendering viewpoint to the viewpoint of the generated images.
This selection process, which we refer to as multi-view score distillation sampling, involves choosing the nearest reference image to the rendered camera view for calculating the SDS loss.

Following the approach \cite{tang2023dreamgaussian}, we utilize the reference image to compute both the reconstruction loss $\mathcal{L}_{rec}$ and the foreground mask loss $\mathcal{L}_{mask}$.
Thus, the final optimization objective is:
\begin{equation}
    \begin{split}
         \mathcal{L} = &\mathcal{L}_{MVSDS} + \lambda_3 \mathcal{L}_{rec} + \lambda_4 \mathcal{L}_{mask},
    \end{split}
\end{equation}
where $\lambda_3$ and $\lambda_4$ are the weighting parameters . 
During training, we first use $\mathcal{L}$ to supervise a fixed frame to get static canonical 3D Gaussian. Then we use all anchors and reference images to train the dynamic 4D Gaussian point cloud.

%\begin{equation}
%    \begin{split}
%         \mathcal{L} = &\mathcal{L}_{MVSDS} + \lambda_3 \mathcal{L}_{rec} + \lambda_4 \mathcal{L}_{mask} + \\
%                  &\lambda_5 \mathcal{L}_{rigid} + \lambda_6 \mathcal{L}_{rot} +\lambda_7 \mathcal{L}_{iso}
%    \end{split}
%\end{equation}

%\subsection{Application: Text-to-4D Generation}
%\begin{figure*}[t]
%  \centering
%  \includegraphics[width=1.0\linewidth]{img/text-to-4d-pipeline.jpg}
%  \vspace{-0.3in}
%  \caption{\textbf{Text to 4D Pipeline.} Combined with image and video diffusion models, our approach could generate 4D content from text. }
%  \vspace{-0.1in}
%\label{fig:text_to_4d_pipline}
%\end{figure*}

\subsection{Text-to-4D Extension}
Our method is designed to be readily adaptable to text and image inputs, offering novel capabilities such as directly generating video sequences from textual descriptions or static images. We initiate this process using a 2D diffusion model SDXL~\cite{podell2023sdxl}, which generates the image from textual input. This 2D output is then transformed into a video sequence via a video diffusion model, such as SVD~\cite{blattmann2023stable}, adding temporal coherence and motion to the static image. Finally, the aforementioned pipeline further lifts the video sequence to a 4D scene, introducing an extra spatial dimension that creates immersive experiences beyond conventional 2D or 3D media. The generation pipeline with the given text, image, and video inputs is visualized in Fig.~\ref{fig:pipeline}.

% Our approach can be easily extended to accept text and image inputs. Notably, the ability to create video sequences directly from text descriptions or static images has opened up exciting possibilities. Specifically, we initiate the process by employing a 2D diffusion model, exemplified by SDXL~\cite{podell2023sdxl}. This model allows us to generate a desired object from textual input. Building upon the created 2D result, we extend it into a video sequence through a video diffusion model, e.g. SVD~\cite{blattmann2023stable}. This introduces temporal coherence and motion variance to the image result.  Finally, our method elevates the video sequence to 4D content. By incorporating an additional spatial dimension, we create immersive results that transcend traditional 2D or 3D media.

\section{Experiment}

% % summary
% In this section, we conduct comprehensive experiments to illustrate the performance of our methods and the effectiveness of each component. 
% We begin by introducing the experimental setting, including the dataset, evaluation metrics, and comparison baselines. 
% Subsequently, we provide detailed implementation specifics of our method. 
% Following this, we present both quantitative and qualitative comparisons with other state-of-the-art methods on both the benchmark and Internet data-set, as well as ablations of each design in our work.

\subsection{Experiment Setup}
\noindent\textbf{Dataset} 
For the video-to-4D task, we utilize the dataset provided by Consistent4D~\cite{jiang2023consistent4d} for quantitative evaluation, which comprises multi-view videos depicting 7 dynamic objects. 
Additionally, for qualitative evaluation, we curate a set of challenging videos from online sources to assess the robustness and generalization capabilities of each method.
%we employ 24 videos sourced from the corresponding GitHub page. Furthermore, \zyf{what does "24 videos sourced from the corresponding GitHub" mean}
For text/image-to-4D tasks, we follow the data settings in 4Dfy~\cite{bahmani20234dfy} and DreamGaussian4D~\cite{ren2023dreamgaussian4d} to create the corresponding results. Notably, we use the same input image and input video with DreamGaussian4D for a fair comparison.

\begin{figure*}[t]
  \centering
  \includegraphics[width=1.0\linewidth]{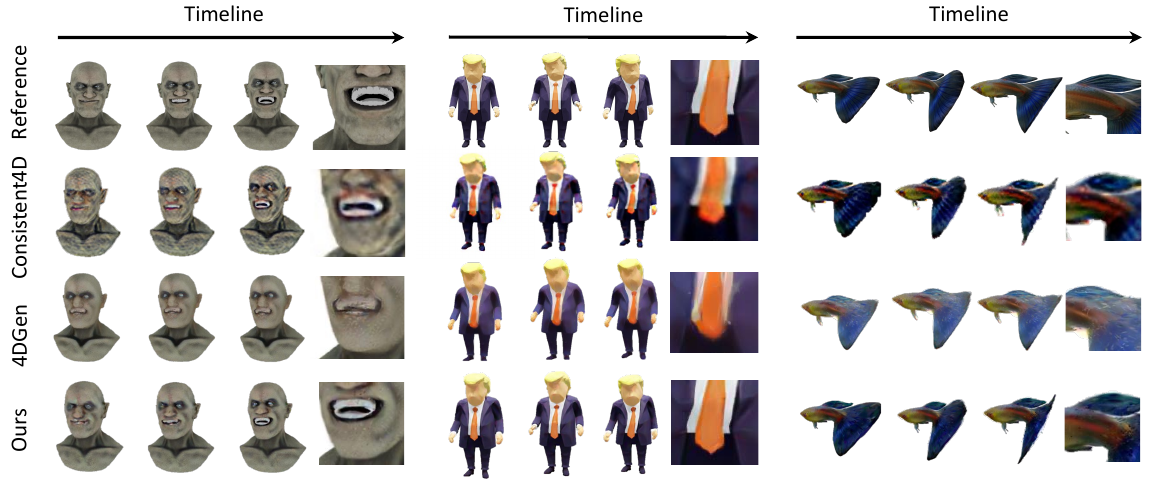}
  \vspace{-0.3in}
  \caption{\textbf{Qualitative comparison on video-to-4D generation}. Our method achieves better faithfulness to the input video with higher quality on the overall reconstruction.}  
  \vspace{-0.1in}
\label{fig:qualtitive_comparision}
\end{figure*}

\noindent\textbf{Evaluation metrics} 
We employ CLIP, LPIPS, and FVD for Video-to-4D evaluation as in the previous approach~\cite{jiang2023consistent4d}. 
Specifically, CLIP and LPIPS serve as image-level metrics, assessing the semantic similarity between rendered images and ground truths; 
FVD is a video-level metric commonly used in video generation tasks, considering not only single-frame quality but also temporal coherence; we also consider the FID-VID metric to measure the temporal consistency.
As 4D-Gen\cite{yin20234dgen} does not support a 30-frame setting we report its FVD under a 16-frame reconstruction, namely FVD-16 as an additional metrics.  
For text-to-4D generation, we follow previous works and provide a user study for quantitative comparisons between different methods. We also provide a user study for video-to-4D and both studies are detailed in the supplementary material. 

\noindent\textbf{Baselines} 
For video-to-4D generation, we compare our method with two baselines: Consistent4D~\cite{jiang2023consistent4d} and the concurrent work 4DGen~\cite{yin20234dgen}. 
We utilize the official code released by the authors to generate comparison results.
%Consistent4D employs a cascade dynamic NeRF to model 4D objects, and we obtain its results from the official GitHub repository. The concurrent work 4DGen adopts 4DGaussian~\cite{wu20234d} as a 4D representation without any modifications. \yifei{do we need to emphasize each method's detail again here?}
For text\&image-to-4D generation, we make comparisons with two state-of-art models that are open-sourced: 4Dfy~\cite{bahmani20234dfy} and DreamGaussian4D~\cite{ren2023dreamgaussian4d}. Their results are generated using codes from their official GitHub repository.

\subsection{Implementation Details}
In the training of the video-to-4D generative model, we adopt a two-stage approach. 
Initially, the model is initialized in canonical space and trained for 1000 steps. 
Subsequently, the deformation fields are learned over 7000 additional steps to accurately capture the dynamic scene. 
For the deformation process, we employ multi-layer perceptrons (MLPs) with 64 hidden layers and 32 hidden features per layer. The initial learning rate for the deformation MLPs is set to $1.6\times 10^{-4}$ and is decayed to $1.6\times 10^{-6}$  by the end of the training. 
In the context of adaptive densification, we opt to densify the top $2.5\%$ of points with the most accumulated gradient.  
% Further details regarding the hyperparameters of the 4D Gaussian model are available in the Supplementary Materials.
In the generative phase, we ultilize the spatial-temporal consistent videos for the multiview score distillation sampling optimization.
Empirically, we set $\lambda$ to 0.5 for our temporally consistent diffusion. 
Regarding the loss functions, we maintain the SDS loss weight at 1 and adjust the weights of other losses accordingly. 
Specifically, the reconstruction loss is assigned to $4\times 10^4$, the mask loss is weighted at $1\times 10^4$. 
The training process requires approximately 1 hour on an RTX 3090 GPU, and the rendering process can be performed at 150 FPS in real time.
For Text\&Image-to-4D, we use SDXL~\cite{podell2023sdxl} for image generation, and apply SVD~\cite{blattmann2023stable} to the generated image to create a corresponding video. Then we use the same setting as Video-to-4D to transfer the video into 4D content.

\begin{figure*}[t]
  \centering
  \includegraphics[width=1.0\linewidth]{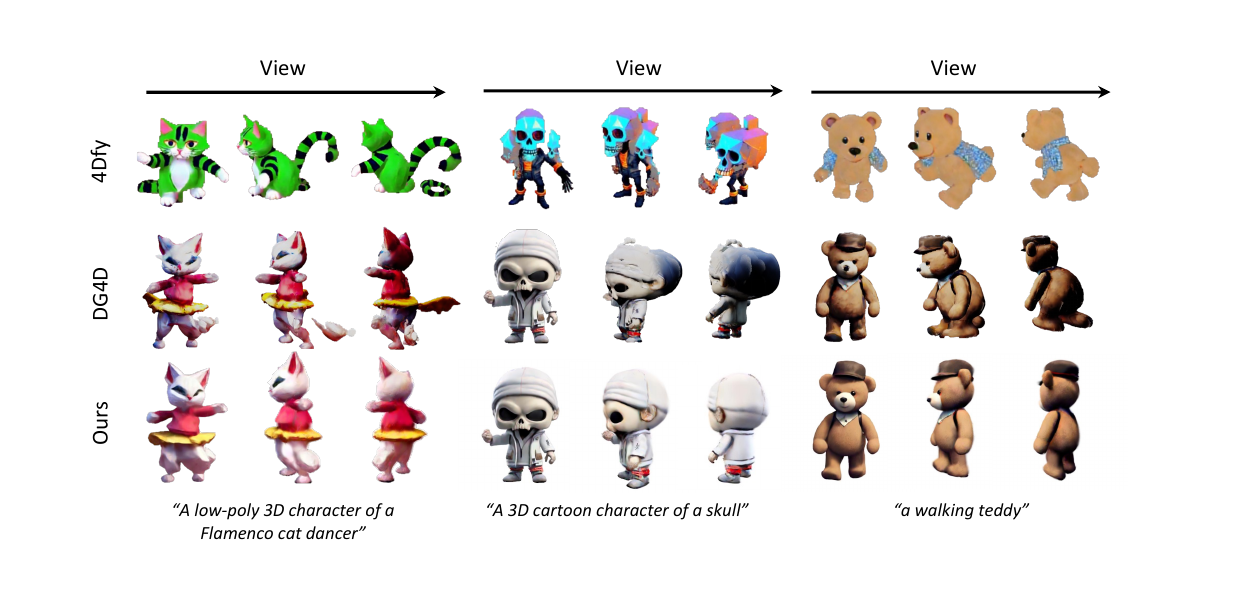}
  \vspace{-0.3in}
  \caption{\textbf{Qualitative comparison on text-to-4D generation}. Our method achieves the best results on the text alignment and visual quality.}  
  \vspace{-0.1in}
\label{fig:qualtitive_comparision_text_to_4d}
\end{figure*}
%ablation_on_recon
%\begin{figure*}[t]
%  \centering
%  \includegraphics[width=1.0\linewidth]{img/ablation_on_recon.jpg}
%  \vspace{-0.3in}
%  \caption{\textbf{Ablation on Zero123++ and TimeMix} We evaluate the effect of Zero123++(Multiview-based SDS) and TimeMix on the reconstruction results. w./o.Zero123++ refers to the case when we use only the front view as the input of SDS, and w./o. TimeMix refers to the case when we directly apply Zero123++ for generating multi-view images without any modification. Results show our approach provides meaningful information on geometry and texture for the reconstruction.}  
%  \vspace{-0.1in}
%\label{fig:ablation_on_recon}
%\end{figure*}
\begin{figure}[tb]
  \centering
    \begin{subfigure}{0.49\linewidth}
    \centering
    \includegraphics[width=1.0\linewidth]{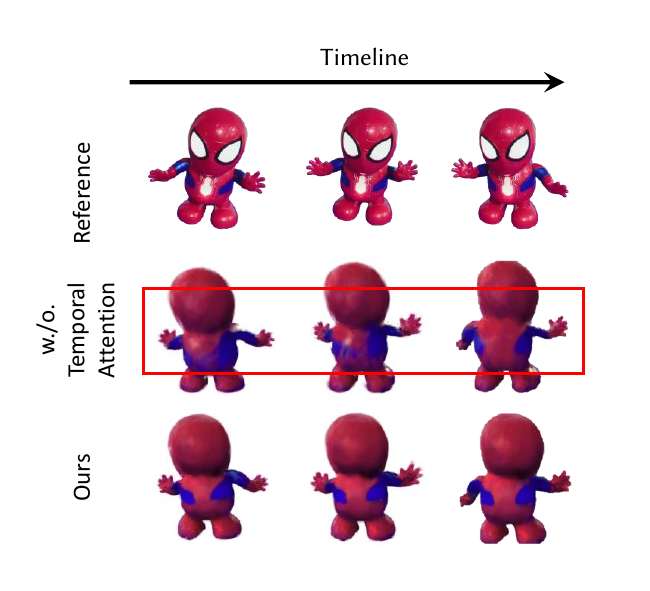}
    %\caption{Another example of a subfigure}
    \label{fig:ablation_on_recon_temporal}
    \end{subfigure}
  \hfill
  \begin{subfigure}{0.49\linewidth}
    \centering
    \includegraphics[width=1.0\linewidth]{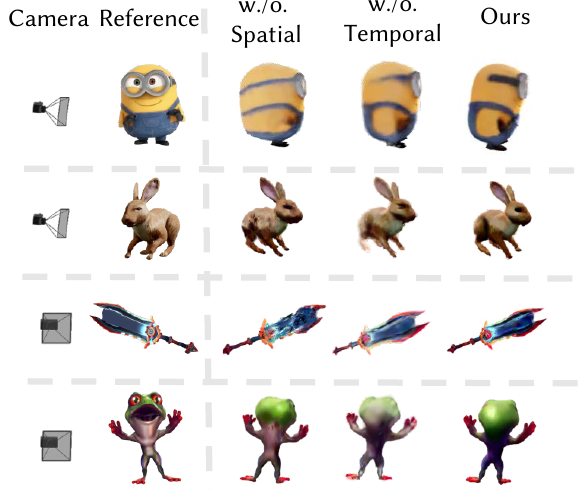}
    %\caption{An example of a subfigure}
    \label{fig:ablation_on_recon_spatial}
  \end{subfigure}
  \caption{\textbf{Ablation on spatial and temporal attention}. We evaluate the effect of spatial and temporal attention on the reconstruction results. 
  %w./o.Zero123++ refers to the case when we use only the front view as the input of SDS, and w./o. TimeMix refers to the case when we directly apply Zero123++ for generating multi-view images without any modification. Results show our approach provides meaningful information on geometry and texture for the reconstruction.
  }
  \label{fig:ablation_recon}

\end{figure}

\subsection{Comparison with State-of-the-art Approaches}
%\jiang{TODO: I didn't check the following paragraphs, please check it!! BTW, "User Study Table 2 shows the user study" seems to have formatting issue, overlapping with the text region.}
In this section, we conduct a comprehensive comparison with the aforementioned baselines using synthetic and in-the-wild data. 
The superior performance demonstrated in both quantitative and qualitative evaluations serves to substantiate the effectiveness and robustness of our proposed method.

% \textbf{Quantitative comparison}
% As shown in Table~\ref{tab:comparison}, the quantitative comparison on video sequences of dynamic scenes captured from multiple viewpoints is provided.
% We compare our method with two state-of-the-art methods~\cite{jiang2023consistent4d,yin20234dgen}. 
% We report three metrics to measure the quality of the generated images and videos: CLIP-similarity 3, LPIPS, and FVD. 
% As shown in Table~\ref{tab:comparison}, our method achieves the best results on all metrics, demonstrating the superior quality and robustness of our method.

\noindent\textbf{Quantitative Results on Video-to-4D}
The experimental results presented in Table~~\ref{tab:comparison} offer a comparative analysis of our method against state-of-the-art techniques, namely Consistent4D and 4DGen, across several metrics. 
Our method outperforms Consistent4D and 4DGen in CLIP, showing better semantic consistency with the target content. It also achieves the lowest LPIPS score, meaning it generates more realistic images than the others. For the video quality and smoothness, our method beats 4DGen in FID-VID and FVD-16, and Consistent4D in FVD, implying that our videos are closer to real videos and have less temporal artifacts. Overall, the numbers confirm that our method is superior in generating semantically aligned and perceptually convincing videos with high fidelity and coherence.

\noindent\textbf{Qualitative Comparison on Video-to-4D} 
In our investigation of the video-to-4D task, we have employed various methodologies to conduct qualitative comparisons.
Figure ~\ref{fig:qualtitive_comparision} shows the rendered outcomes at distinct temporal intervals and perspectives.
A critical observation of our findings reveals that Gaussian splatting is inadequate in producing credible and temporally consistent images during per-frame reconstruction.
Furthermore, we have observed that Consistent4D often produces over-saturated and unrealistic patterns, which may stem from the over-restrictive geometry representation in their cascade Dynerf.
Additionally, our analysis indicates that 4DGen struggles to generate reasonable motion for the Gaussians, leading to its points lacking deformation over time. 
Moreover, 4DGen fails to generate detailed textures for the surface, which appear to be over-smoothed in some cases.
Conversely, our methodology demonstrates substantial enhancements in comparison to existing strategies, particularly in aspects of reconstruction fidelity and stability.
Notably, our technique facilitates seamless color transitions, as exemplified in the Fish case, and renders a convincing low-poly effect, as demonstrated in the Trump case, underscoring our method’s adaptability to diverse dynamic environments.

\noindent\textbf{Qualitative Comparison on Text-to-4D} We also evaluate our method on the Text\&Image-to-4D task, which aims to generate 4D content from text and image inputs. Figure~\ref{fig:qualtitive_comparision_text_to_4d} shows the 4D generation results of our method and two baselines, 4Dfy~\cite{bahmani20234dfy} and DreamGaussian4D~\cite{ren2023dreamgaussian4d}, under different views. As can be seen, 4Dfy~\cite{bahmani20234dfy} produces coarse results and fails to capture the complex semantics of the text input, e.g. in the Cat case. DreamGaussian4D~\cite{ren2023dreamgaussian4d} generates meshes with fine-grained details in the front view, but suffers from severe artifacts and distortions in the side view or back view. In contrast, our method synthesizes Gaussians with smooth and consistent geometry and realistic texture across different views, thanks to our attention mechanism that leverages spatial and temporal information. Our method demonstrates its superiority and versatility on the Text\&Image-to-4D task, which can be seen as a direct application of our Video-to-4D framework.
\vspace{-0.5cm}
%\input{tab/user_study}
%\begin{table}[ht]
%  \centering
%  \begin{tabular}{lccccc}
%    \toprule
%     & CLIP $\uparrow$ & LPIPS $\downarrow$ & FID-VID $\downarrow$ & FVD $\downarrow$ & FVD-16 $\downarrow$\\
%    \hline
%    Consistent4D &0.877 &0.134 & / &1133.93 &/\\
%    4DGen & 0.894& 0.130& 71.99& /&1005.72\\
%    Ours & 0.909 &0.126 &52.58 & 992.21 &952.41\\
%  \bottomrule
%  \end{tabular}
%  \caption{Comparison with state-of-the-art methods.}
%  \label{tab:comparison}
%\end{table}
\begin{table*}[ht]
  \centering
  \vspace{-0.36cm}
  % First minipage for the first table
  \begin{minipage}[t]{0.62\textwidth}
    \centering
    \caption{Comparison with state-of-the-art methods.}
    \begin{adjustbox}{width=\textwidth}

          \centering
          \begin{tabular}{lccccc}
            \toprule
             & CLIP $\uparrow$ & LPIPS $\downarrow$ & FID-VID $\downarrow$ & FVD $\downarrow$ & FVD-16 $\downarrow$\\
            \hline
            Consistent4D &0.877 &0.134 & / &1133.93 &/\\
            4DGen & 0.894& 0.130& 71.99& /&1005.72\\
            Ours & \textbf{0.909} &\textbf{0.126} &\textbf{52.58} & \textbf{992.21} &\textbf{952.41}\\
          \bottomrule
          \end{tabular}

    \end{adjustbox}

    \label{tab:comparison}
  \end{minipage}
  \hfill
  % \begin{minipage}[t]{0.35\textwidth}
  %   \centering
  %   \caption{User study on best Video-to-4D methods.}
  %   \begin{adjustbox}{width=\textwidth}

  %           \centering
  %           \vspace{-0.1in}
  %           \begin{tabular}{@{}lccc@{}}
  %           \toprule
  %                       & Vis. & Cons. & Align.   \\
  %           \hline
  %           Consistent4D & 28.6\% & 28.6\% & 35.7\%    \\
  %           4DGen & 0\% & 0\% & 0\%  \\
  %           Ours        & \textbf{71.4}\% & \textbf{71.4}\% & \textbf{64.3}\% \\ 
  %           \bottomrule
  %           \end{tabular}
  %   \end{adjustbox}

  %   \label{tab:user_study}
  % \end{minipage}
\end{table*}

%User study on the best-performing 4D-to-Video generation methods. We ask respondents to compare the three methods on visual quality, temporal consistency, and generation consistency to input video. Totally 31 respondents were invited and the proposed method demonstrated clear popularity compared to the alternatives.

% \textbf{User Study} 
% We conducted a user study on Video-to-4D generation tasks, where we compared our method with Consistent4D and 4DGen on 14 test cases. We asked the participants to rate the methods on visual quality (Vis.), temporal consistency (Cons.), and alignment with input videos (Align.). Table~\ref{tab:user_study} shows that 4DGen got no votes in any category, and Consistent4D got 28.6\% for Vis. and Cons., and 35.7\% for Align. Our method was preferred by most participants in all categories, with 71.4\% votes for Vis. and Cons., and 64.3\% for Align. These results demonstrate that our method outperforms the others in generating high-quality and coherent Video-to-4D content.
\vspace{-0.5cm}
\subsection{Ablation Study}

The experimental analysis presented in Table~\ref{tab:ablation} provides a comprehensive evaluation of the impact of various components on the performance of a diffusion model with a 4D representation. 
The components under consideration include baseline diffusion, spatial anchor, spatial-temporal anchor, and adaptive densification.
In the first configuration of Table~\ref{tab:ablation}, we regard the model utilizing the diffusion process~\cite{liu2023zero} with adaptive densification as the baseline.
\begin{table*}[ht]
  \centering

  % First minipage for the first table
  \begin{minipage}[t]{0.57\textwidth}
    \centering
    \caption{Ablation study on effectiveness of each novel component.}
    \begin{adjustbox}{width=\textwidth}
    \begin{tabular}{@{}cccc|cccc@{}}
      \toprule
      \multicolumn{3}{c}{Diffusion Model} & \multicolumn{1}{c|}{4D Gaussian} & \multicolumn{4}{c}{Evaluation Metrics} \\
      Baseline & Spat. & Temp. & Adaptive Dens. & CLIP $\uparrow$ & LPIPS $\downarrow$ & FID-VID $\downarrow$ & FVD $\downarrow$\\
      \midrule
      \checkmark & & & \checkmark &0.895 & 0.135&77.88 & 1369.65\\
      \checkmark & \checkmark & & \checkmark & 0.899&0.130 &67.04 &1198.23\\
      \checkmark & \checkmark & \checkmark & & 0.890 &0.136 &90.30 &1453.53 \\
      \checkmark & \checkmark &\checkmark& \checkmark &\textbf{0.909} &\textbf{0.126} &\textbf{52.58} & \textbf{992.21}\\
      \bottomrule
    \end{tabular}
    \end{adjustbox}

    \label{tab:ablation}
  \end{minipage}
  \hfill
  \begin{minipage}[t]{0.4\textwidth}
    \centering
    \caption{Ablation study on different attention mechanisms.}
    \begin{adjustbox}{width=\textwidth}
    \begin{tabular}{@{}lcccc@{}}
      \toprule
      & \multicolumn{4}{c}{Evaluation Metrics} \\
       & CLIP $\uparrow$ & LPIPS $\downarrow$ & FID-VID $\downarrow$ & FVD $\downarrow$ \\
      \midrule
      Spat. only &0.899 & 0.130& 67.04&1198.23 \\
      Cross-Frame & 0.901& 0.129& 63.32&1053.25 \\
      Spat.-Temp. &\textbf{0.909} & \textbf{0.126}& \textbf{52.58}&  \textbf{992.21}\\
      \bottomrule
    \end{tabular}
    \end{adjustbox}
    \vspace{0.075cm}

    \label{tab:temporal_attn}
  \end{minipage}
\end{table*}

%\begin{table}[ht]
%  \centering
%
%  % First minipage for the first table
%
%
%    \begin{tabular}{@{}cccc|cccc@{}}
%      \toprule
%      \multicolumn{3}{c}{Diffusion Model} & \multicolumn{1}{c|}{4D Gaussian} & \multicolumn{4}{c}{Evaluation Metrics} \\
%      Baseline & Spatial & Temporal & Adaptive Dens. & CLIP $\uparrow$ & LPIPS $\downarrow$ & FID-VID $\downarrow$ & FVD $\downarrow$\\
%      \midrule
%      \checkmark & & & \checkmark &0.895 & 0.135&77.88 & 1369.65\\
%      \checkmark & \checkmark & & \checkmark & 0.899&0.130 &67.04 &1198.23\\
%      \checkmark & \checkmark & \checkmark & & 0.890 &0.136 &90.30 &1453.53 \\
%      \checkmark & \checkmark &\checkmark& \checkmark &\textbf{0.909} &\textbf{0.126} &\textbf{52.58} & \textbf{992.21}\\
%      \bottomrule
%    \end{tabular}
%
%    \caption{Ablation study on each novel component.}
%    \label{tab:ablation}
%
%\end{table}
%
%\begin{table}[ht]
%  \centering
%
%
%    \begin{tabular}{@{}lcccc@{}}
%      \toprule
%      & \multicolumn{4}{c}{Evaluation Metrics} \\
%       & CLIP $\uparrow$ & LPIPS $\downarrow$ & FID-VID $\downarrow$ & FVD $\downarrow$ \\
%      \midrule
%      Spatial only &0.899 & 0.130& 67.04&1198.23 \\
%      Cross-Frame & 0.901& 0.129& 63.32&1053.25 \\
%      Spatial-Temporal &\textbf{0.909} & \textbf{0.126}& \textbf{52.58}&  \textbf{992.21}\\
%      \bottomrule
%    \end{tabular}
%
%    \vspace{0.075cm}
%    \caption{Ablation study on different attention mechanisms.}
%    \label{tab:temporal_attn}
%\end{table}
\noindent\textbf{Spatial and Temporal Anchors} 
Table~\ref{tab:ablation} shows that adding the spatial anchor to the baseline model leads to a marginal improvement in all metrics, especially FID-VID and FVD. The final model, which uses both spatial-temporal anchor and adaptive densification, improves all metrics significantly. It has the best CLIP, LPIPS, FID-VID, and FVD scores, meaning it generates the most semantically consistent, realistic, and smooth videos. 

Figure~\ref{fig:ablation_recon} shows the importance of our spatial-temporal anchor for improving the 4D content quality. The baseline 4D Gaussian method, labeled as “w./o. spatial,” fails to deform the point clouds properly, leading to unreasonable shapes without spatial-temporal constraints. A model that uses only the spatial anchor, “w./o. temporal,” struggles to deform solid shapes, as seen in the rabbit’s leg and the frog’s back. Our method, which combines both spatial and temporal anchors, overcomes these issues and produces visually consistent and accurate 4D content.

\begin{figure}[tb]
  \centering
  \begin{subfigure}{0.49\linewidth}
      \centering
      \includegraphics[width=1.0\linewidth]{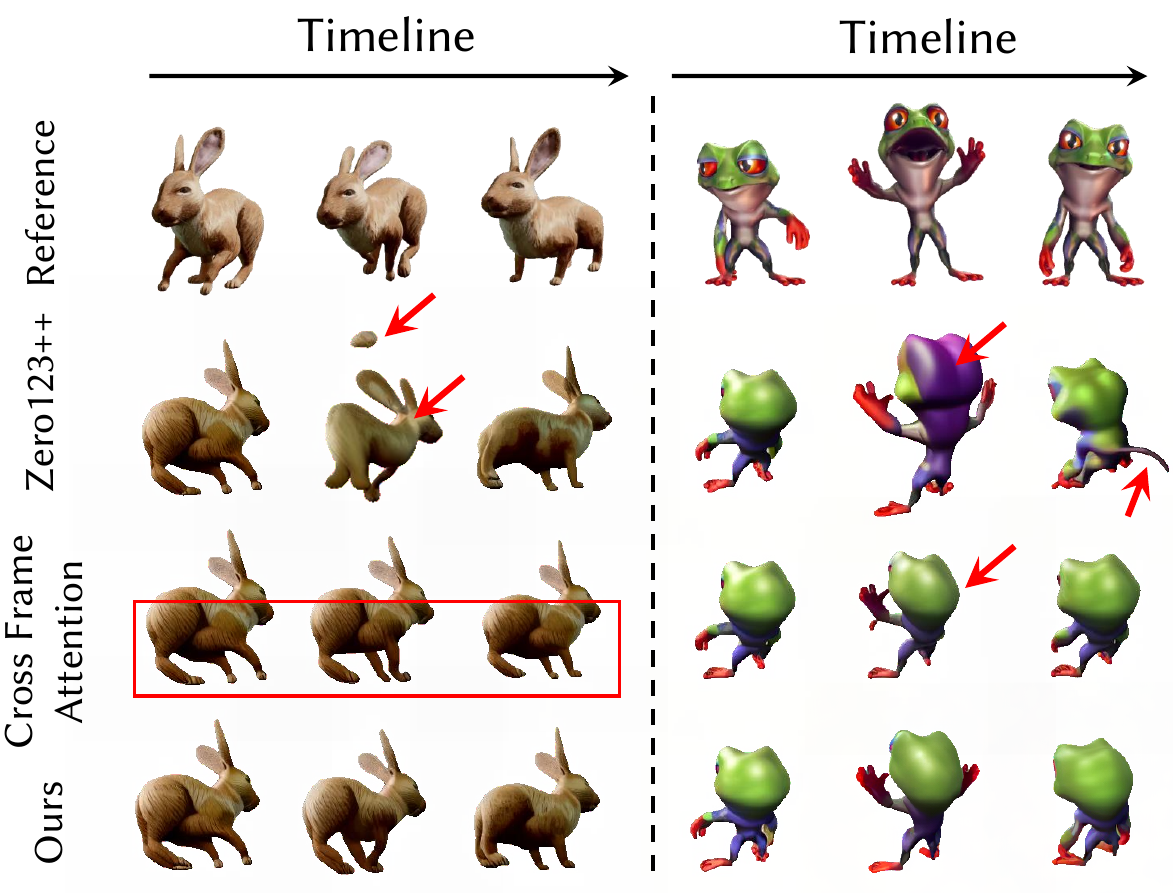}
      \caption{\textbf{Ablation on Multiview Diffusion Models} We evaluate the effect of different attention mechanism on Zero123++. Our attention mechanism can achieve both temporal coherence and geometric fidelity given a monocular input video.}  
    \label{fig:ablation_on_diffusion}
  \end{subfigure}
  \hfill
  \begin{subfigure}{0.49\linewidth}
      \centering
      \includegraphics[width=1.0\linewidth]{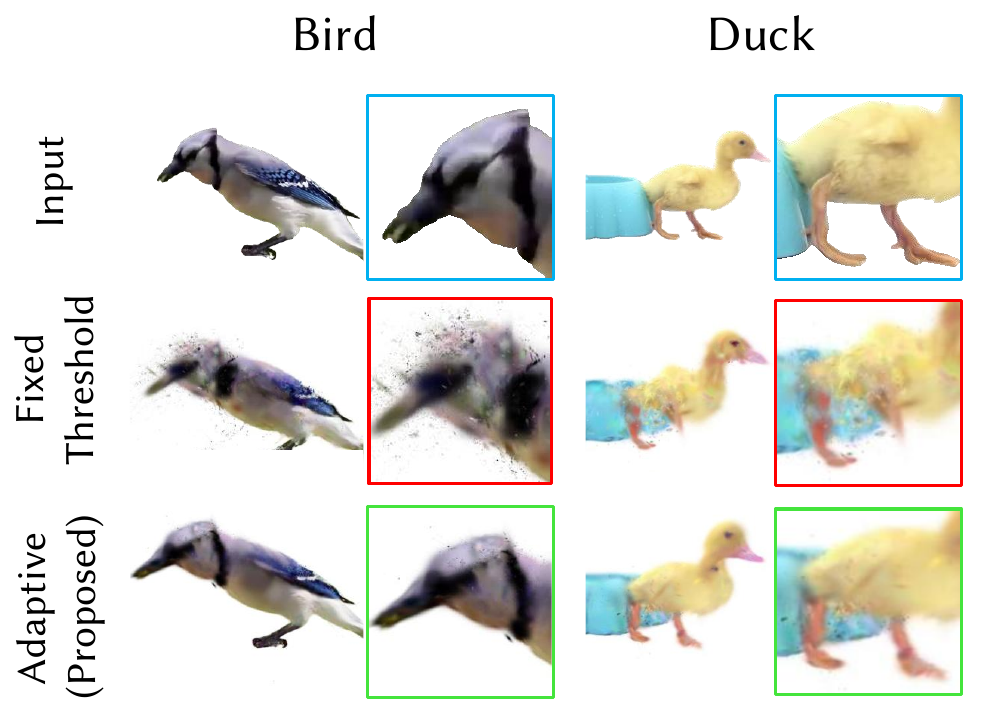}
      \caption{\textbf{Ablation on Adaptive Densification} We evaluate the effect of using different densify mechanism during reconstruction. Our adaptive threshold provides reasonable densification guidance for each case.}  
    \label{fig:ablation_on_densify}
  \end{subfigure}
\end{figure}

\noindent\textbf{Adaptive Densification} We present the statistical analysis of the adaptive densification module in the third and final configurations of Table~\ref{tab:ablation}. The results show that the adaptive densification module improves the overall quality of the generative model, especially in terms of the FID-VID and FVD metrics. We also compare the proposed adaptive densification with a fixed threshold approach in Figure~\ref{fig:ablation_on_densify}. We observe that a fixed threshold can lead to either under-densification or over-densification, depending on the case. For example, in the Bird case, the fixed threshold is too low for densification, which results in excessive Gaussian points. On the other hand, in the Duck case, the fixed threshold is too high for densification, which causes the foot to be blurry due to insufficient Gaussian points. Our method overcomes this limitation by adapting the gradient threshold to different cases, thus producing robust and stable 4D Gaussian results.

%\begin{figure}[t]
%  \centering
%  \includegraphics[width=1.0\linewidth]{img/ablation_on_diffusion.pdf}
%  \caption{\textbf{Ablation on Multiview Diffusion Models} We evaluate the effect of different attention mechanism on Zero123++. Our TimeMix mechanism can achieve both temporal coherence and geometric fidelity given a monocular input video.}  
%\label{fig:ablation_on_diffusion}
%\end{figure}
%
%\begin{figure}[t]
%  \centering
%  \includegraphics[width=0.45\linewidth]{img/ablation_on_densify.pdf}
%  \caption{\textbf{Ablation on Adaptive Densification}. We evaluate the effect of using different densify mechanism during reconstruction. Our adaptive threshold provides reasonable densification guidance for each case.}  
%\label{fig:ablation_on_densify}
%\end{figure}

\textbf{Different Attention Mechanisms}
In our study, we evaluated three distinct attention mechanisms within the context of images generated by our diffusion model: spatial attention, cross-frame attention, and spatial-temporal attention. The visual outcomes of these varied approaches are illustrated in Figure~\ref{fig:ablation_on_diffusion}. We employed a parameter $\lambda$ to modulate the extent of spatial and temporal attention exerted, as delineated in Equation~\ref{eq:spatial_temporal}.
Our findings indicate that exclusive reliance on spatial attention does not guarantee consistency across sequential frames. While cross-frame attention enhances frame-to-frame coherence, it fails to achieve a satisfactory alignment with reference images. Conversely, the integration of spatial-temporal attention ensures consistency across both dimensions.
We quantitatively assessed the quality of images generated under different attention regimes using four established metrics: CLIP, LPIPS, FID-VID, and FVD. The comparative results are tabulated in Table~\ref{tab:temporal_attn}. It was observed that images generated with spatial-temporal attention outperformed others across all evaluation metrics. This underscores the efficacy of our proposed attention mechanism in augmenting the diffusion model for enhanced 4D content generation.

\subsection{Limitations}
While our method has shown promising results, it is important to acknowledge several limitations. 
Specifically, our approach is constrained in handling complicated and fast motion due to the inherent limitations of current 4D representation in terms of Gaussian Splatting, which can lead to suboptimal outcomes. 
Additionally, inherent video limitations such as image blurriness can restrict the effectiveness of diffusion, consequently impacting the subsequent optimization of 4D Gaussian Splatting. 
Furthermore, domain transfer to videos lacking well-segmented foreground or containing multiple foreground objects presents a non-trivial challenge that warrants consideration. 
These limitations are important to address in future research and development efforts.

\section{Conclusion}
The paper presents a novel approach for dynamic 3D content generation from monocular videos, addressing the challenges of 4D representation and spatial-temporal consistency. 
By leveraging specially tailored 4D Gaussian splatting and a novel information fusion module, the proposed method achieves high-quality and robust 4D scene generation.  
Comprehensive experiments demonstrate the method's effectiveness, showcasing an obvious faster generation speed and significant improvements in rendering quality and temporal consistency compared to prior state-of-the-art methods. 
Overall, the proposed method sets a new benchmark for training speed, rendering quality, and 4D consistency in dynamic 3D content generation from monocular videos, opening up possibilities for real-world applications.

% \par\vfill\par
% Now we have reached the maximum length of an ECCV \ECCVyear{2024} submission (excluding references).
% References should start immediately after the main text, but can continue past p.\ 14 if needed.
\clearpage  % TODO REVIEW/FINAL: This \clearpage needs to be removed from both review and camera-ready versions.

% ---- Bibliography ----
%
% BibTeX users should specify bibliography style 'splncs04'.
% References will then be sorted and formatted in the correct style.
%
\bibliographystyle{splncs04}
\bibliography{main}

% ---------------------------------------------------------------
% TODO REVIEW: Replace with your title
\title{Supplementary Materials} 

% TODO REVIEW: If the paper title is too long for the running head, you can set
% an abbreviated paper title here. If not, comment out.
\titlerunning{STAG4D}

% TODO FINAL: Replace with your author list. 
% Include the authors' OCRID for the camera-ready version, if at all possible.
\author{}
%% TODO FINAL: Replace with an abbreviated list of authors.
\authorrunning{Yifei Zeng et al.}
% First names are abbreviated in the running head.
% If there are more than two authors, 'et al.' is used.

% TODO FINAL: Replace with your institution list.
\institute{}
%Springer Heidelberg, Tiergartenstr.~17, 69121 Heidelberg, Germany
%\email{lncs@springer.com}\\
%\url{http://www.springer.com/gp/computer-science/lncs} \and
%ABC Institute, Rupert-Karls-University Heidelberg, Heidelberg, Germany\\
%\email{\{abc,lncs\}@uni-heidelberg.de}}

\maketitle

%\begin{abstract}
%  The abstract should concisely summarize the contents of the paper. 
%  While there is no fixed length restriction for the abstract, it is recommended to limit your abstract to approximately 150 words.
%  Please include keywords as in the example below. 
%  This is required for papers in LNCS proceedings.
%  \keywords{First keyword \and Second keyword \and Third keyword}
%\end{abstract}

\section{Graident Distribution Analysis for Adaptive Densification}
\begin{figure}[h]
  \centering
  \includegraphics[width=0.5\linewidth]{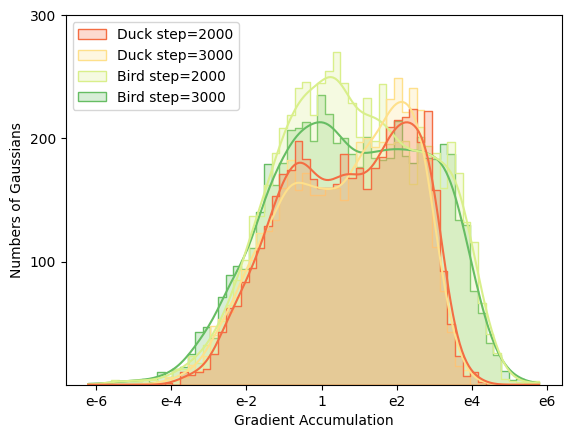}
  \caption{Gradient Distribution Analysis for Adaptive Densification
  }
  \label{fig:gradient_visualize}
\end{figure}
In Fig~\ref{fig:gradient_visualize}, we present a visual examination of the gradient distribution for the Duck and Bird scenarios across various training epochs. This analysis elucidates the rationale behind the ablation study of our adaptive densification approach. The graphical data indicates that the gradient distributions for different scenarios converge to disparate values, signifying a substantial numerical disparity when plotted on a logarithmic scale. For the Duck scenario, a suitable gradient threshold is approximately 40 ($e^{3.7}$), whereas for the Bird scenario, it is around 150 ($e^5$). Moreover, the gradient distribution maintains a consistent shape throughout the training phase for each case. Consequently, we introduce the adaptive densification mechanism that dynamically modulates the threshold in response to the specific scenario, which is to densify the Gaussian points that exhibit relatively higher gradient accumulation. This strategy is anticipated to enhance the robustness and quality of the generative performance.

\section{User Study}

\subsection{Video-to-4D}

In the conducted user study focusing on Video-to-4D content generation, we juxtaposed our methodology against Consistent4D\cite{jiang2023consistent4d} and 4DGen\cite{yin20234dgen} across 14 test instances. Participants were solicited to appraise the methods according to visual quality (Vis.), temporal consistency (Cons.), and alignment with the input videos (Align.). As illustrated in Table~\ref{tab:user_study_video}, 4DGen did not receive any endorsements in the evaluated categories. In contrast, Consistent4D was acknowledged with 28.6\% for Vis. and Cons., and 35.7\% for Align. Predominantly, our method was favored, securing 71.4\% of the votes for Vis. and Cons., and 64.3\% for Align. These outcomes attest to the superior efficacy of our method in synthesizing high-fidelity and temporally coherent Video-to-4D content.
\begin{table}[]
\centering
\vspace{-0.1in}
\begin{tabular}{@{}lccc@{}}
\toprule
            & Vis. & Cons. & Align.   \\
\hline
Consistent4D & 28.6\% & 28.6\% & 35.7\%    \\
4DGen & 0\% & 0\% & 0\%  \\
Ours        & \textbf{71.4\%} & \textbf{71.4}\% & \textbf{64.3}\% \\ 
\bottomrule
\end{tabular}
\caption{User study on the best-performing Video-to-4D generation methods.}
\label{tab:user_study_video}
\end{table}

\vspace{-1.2cm}
\subsection{Text/Video-to-4D}

A user study was conducted to evaluate the efficacy of Text\&Video-to-4D content generation. For this purpose, 4Dfy\cite{bahmani20234dfy} and DreamGaussian4D\cite{ren2023dreamgaussian4d} were selected as comparative benchmarks. The study encompassed 14 test scenarios, and 30 evaluators were recruited to assess the methods based on visual quality (Vis.), temporal consistency (Cons.), and congruence with the input text (Align.). As delineated in Table ~\ref{tab:user_study_text}, our approach garnered the highest ratings across all metrics. These findings underscore our method's preeminence and adaptability in the Text\&Image-to-4D content generation domain.

\begin{table}[]
\centering
\vspace{-0.1in}
\begin{tabular}{@{}lccc@{}}
\toprule
            & Vis. & Cons. & Align.   \\
\hline
4Dfy & 2.04 & 2.15 &  2.26  \\
DG4D & 2.91 & 2.92 & 3.26  \\
Ours        & \textbf{4.02} & \textbf{4.41} & \textbf{4.15} \\ 
\bottomrule
\end{tabular}
\caption{User study for Text\&Image-to-4D generation methods.}
\label{tab:user_study_text}
\end{table}

\vspace{-1.2cm}
\section{Attention Mechanism for Normal Map}
\begin{figure}[tb]
  \centering
  \includegraphics[width=1.0\linewidth]{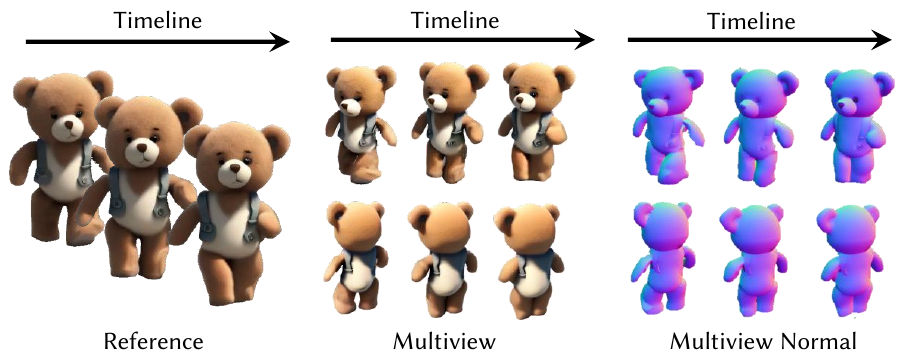}
  \caption{\textbf{Normal Map Illutstration}
  }
  \label{fig:normal_illustration}
\end{figure}
Building upon our attention mechanism, we have integrated our design with the newly introduced normal map model by Zero123++\cite{shi2023zero123++}. This adaptation enables the generation of coherent multiview images with corresponding normal maps, as depicted in Figure ~\ref{fig:normal_illustration}. The successful application of our method to normal maps exemplifies the flexibility and applicability of our attention framework. This enhancement not only broadens the scope of our model but also demonstrates its potential for diverse 4D content generation scenarios.

\section{More Results for Multiview Generation}

\begin{figure}[tb]
    \vspace{-1cm}
  \centering
  \includegraphics[width=1.0\linewidth]{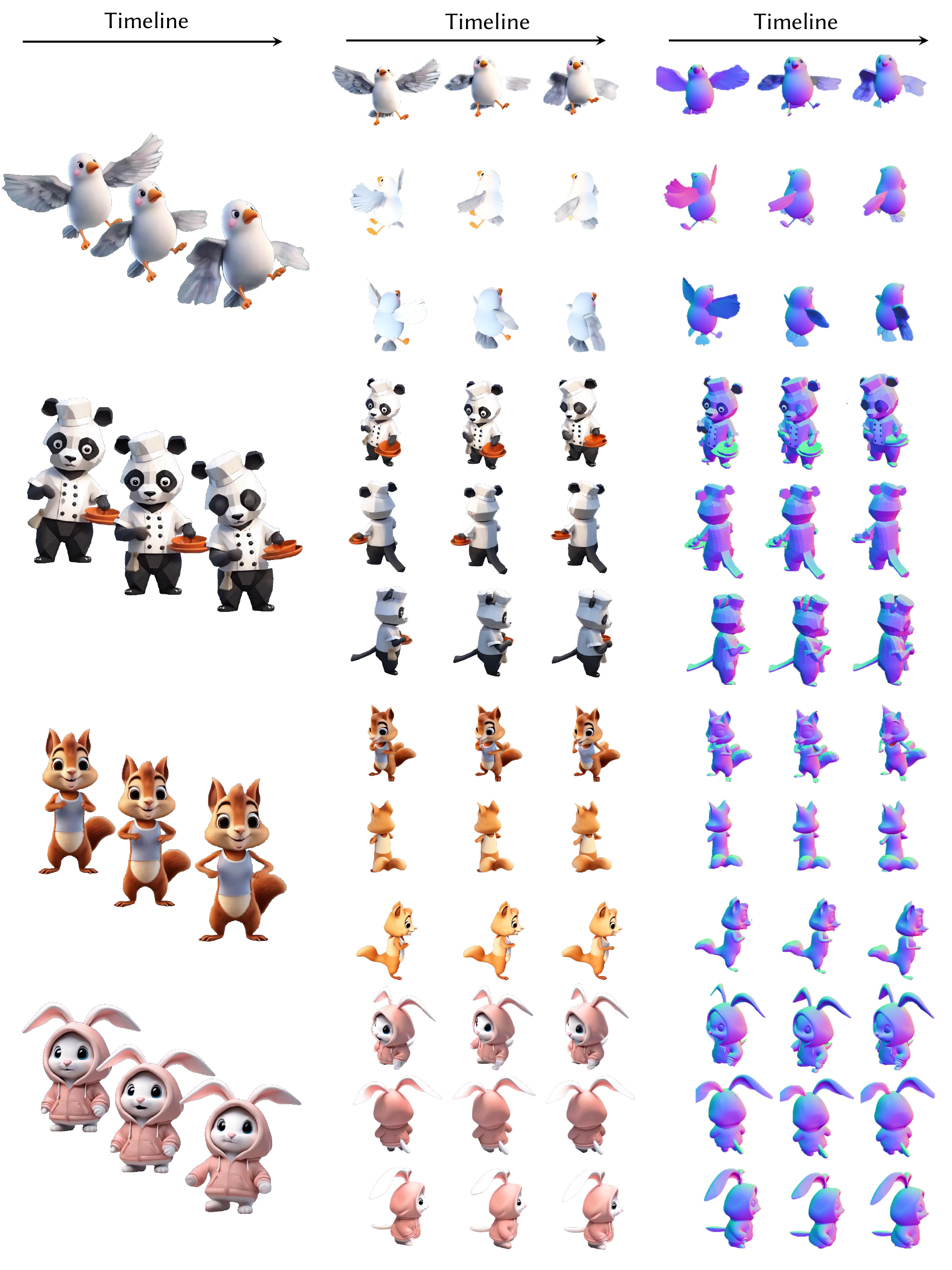}
  \caption{More Results for Multiview Generation (Text input)
  }
  \label{fig:mv_results}
\end{figure}
\begin{figure}[tb]
    \vspace{-1cm}
  \centering
  \includegraphics[width=1.0\linewidth]{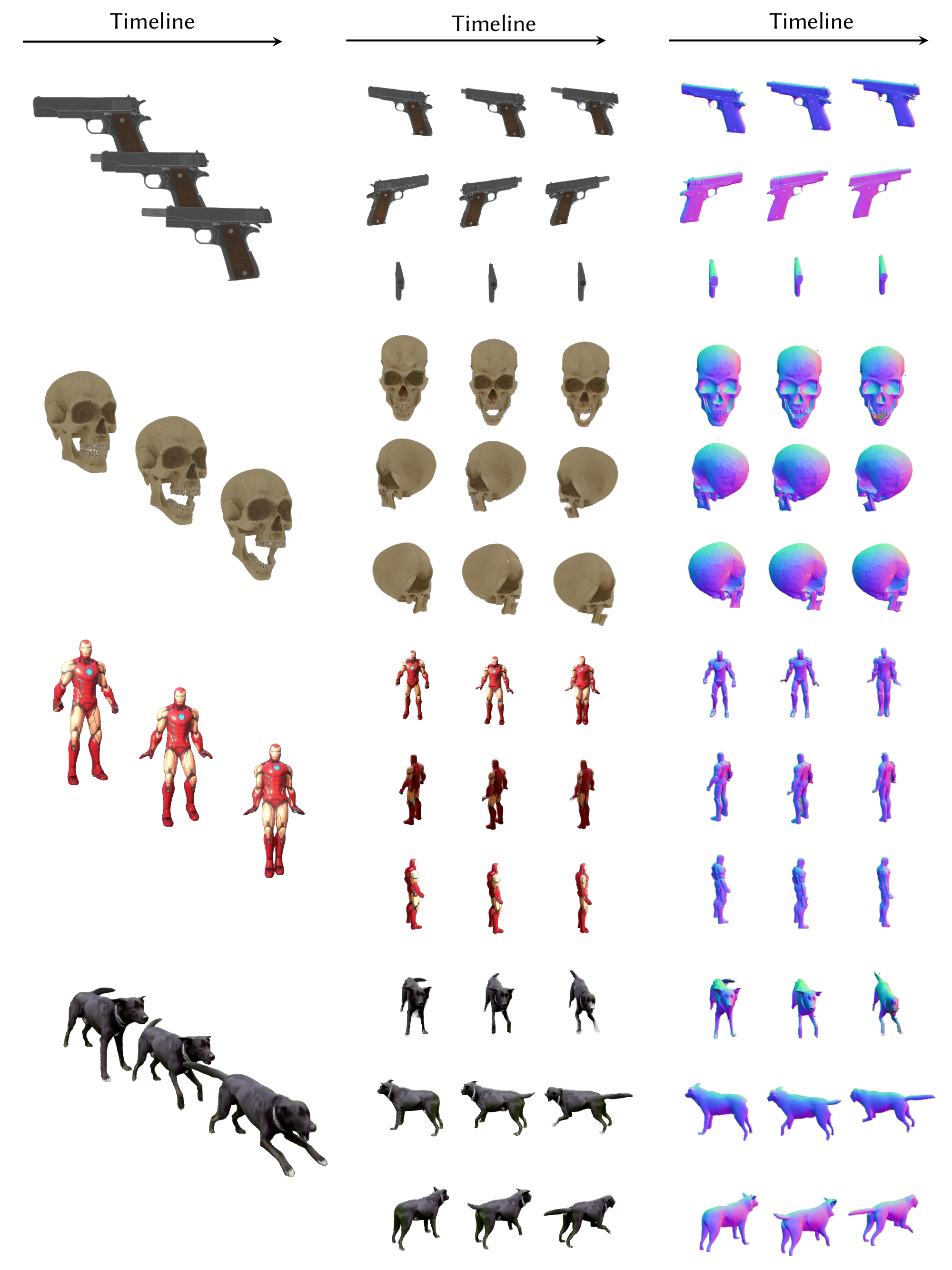}
  \caption{More Results for Multiview Generation (Video input)
  }
  \label{fig:mv_results2}
\end{figure}
We show other results of our spatial and temporal consistent multiview images in Figure ~\ref{fig:mv_results} \& Figure ~\ref{fig:mv_results2}. These results demonstrate the robustness and effectiveness of our method.

\section{More Results for 4D Generation}
\begin{figure}[tb]
    \vspace{-1cm}
  \centering
  \includegraphics[width=1.0\linewidth]{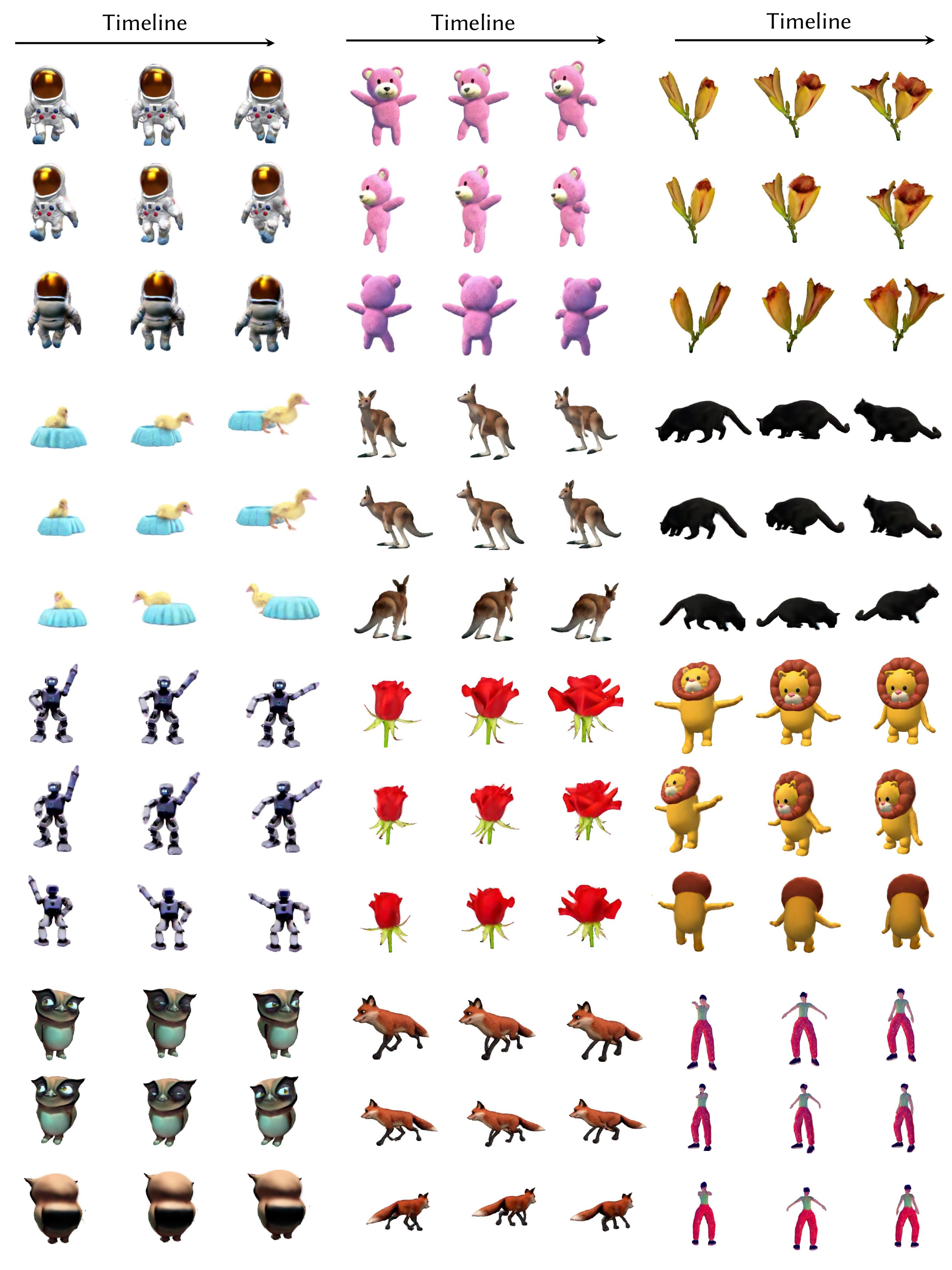}
  \caption{More Results for 4D Generation
  }
  \label{fig:gen_result}
\end{figure}

We show more results of our 4D Generation in Figure ~\ref{fig:gen_result}. Our method could handle various objects and motions, which applies widely to different cases.

\section{Discussion on the Potential Negative Impact}
The generation process of our method involves various pre-trained diffusion models, which still have a controversy on the copyright of the generated result. This will remain an ethical issue until the relevant laws are matured.

% ---- Bibliography ----
%
% BibTeX users should specify bibliography style 'splncs04'.
% References will then be sorted and formatted in the correct style.
%

\end{document}